\documentclass{article}

\usepackage{microtype}
\usepackage{graphicx}
\usepackage{subcaption}
\usepackage{booktabs} 
\usepackage{xspace}
\usepackage{hyperref}

\newcommand{\circled}[1]{\ding{\numexpr#1+201}}
\usepackage{tikz}
\newcommand\encircle[1]{%
  \tikz[baseline=(X.base)] 
    \node (X) [draw, shape=circle, inner sep=-2, fill=black, text=white] {\strut\footnotesize #1};%
}

\newcommand{\eg}{e.g.,\xspace}
\newcommand{\ie}{i.e.,\xspace}

\newcommand{\xname}{EdgeRAG\xspace}

\newcommand\thicktilde{{\lower.74ex\hbox{\texttt{\char`\~}}}}

\newlength{\figcapspc}
\newlength{\figsubcapspc}
\newcommand{\cmark}{\textcolor{green!80!black}{\ding{51}}}%
\newcommand{\xmark}{\textcolor{red!80!black}{\ding{55}}}%

\usepackage{soul}
\usepackage{mdframed}
\mdfdefinestyle{revstyle}{linecolor=yellow, linewidth=2pt, innertopmargin=2pt, innerbottommargin=2pt, innerleftmargin=2pt, innerrightmargin=2pt}

\soulregister\cite7
\soulregister\ref7
\soulregister\pageref7

\makeatletter
\def\SOUL@hlpreamble{%
\setul{}{2.2ex}
\let\SOUL@stcolor\SOUL@hlcolor
\SOUL@stpreamble
}
\makeatother
\soulregister\xname7
\soulregister\cite7
\soulregister\ref7
\soulregister\eg7
\soulregister\ie7
\soulregister\texttt7
\soulregister\circled7
\soulregister\encircle7
\soulregister\uline7
\usepackage{multirow}
\usepackage{enumerate}
\usepackage{enumitem}

\usepackage[accepted]{icml2025}

\usepackage{amsmath}
\usepackage{amssymb}
\usepackage{mathtools}
\usepackage{amsthm}
\usepackage{pifont}

\icmltitlerunning{\xname{}: Online-Indexed RAG For Edge Devices}

\usepackage{authblk}

\makeatletter
\renewcommand\AB@affilsepx{, \protect\Affilfont}
\makeatother

\title{\bfseries \xname{}: Online-Indexed RAG for Edge Devices}
\author[1]{Korakit Seemakhupt}
\author[2]{Sihang Liu}
\author[1,3]{Samira Khan}
\affil[1]{ University of Virginia} 
\affil[2]{ University of Waterloo}
\affil[3]{ Google}
\date{}
\begin{document}
\twocolumn[
\maketitle



\icmlsetsymbol{equal}{*}






]

\begin{abstract}
Deploying Retrieval Augmented Generation (RAG) on resource-constrained edge devices is challenging due to limited memory and processing power.
In this work, we propose \xname{} which addresses the memory constraint by pruning embeddings within clusters and generating embeddings on-demand during retrieval. To avoid the latency of generating embeddings for large tail clusters, \xname{} pre-computes and stores embeddings for these clusters, while adaptively caching remaining embeddings to minimize redundant computations and further optimize latency.
The result from BEIR suite shows that \xname{} offers significant latency reduction over the baseline IVF index, but with similar generation quality while allowing all of our evaluated datasets to fit into the memory.
\end{abstract}

\section{Introduction}

Large Language Models (LLMs) enable new applications such as smart assistants \cite{gassistant, copilot}.
The processing of these powerful LLMs is usually offloaded to the datacenter due to the enormous resources required. 
However, the latest mobile platforms enable smaller LLM to run locally. These lightweight models cannot directly compare with the state-of-the-art hundred-billion parameter LLMs. To enhance these models to process users' custom data and applications, 
a promising solution is to build a compounding system. 
By integrating LLMs with Retrieval Augmented Generation (RAG), these smaller models can leverage local personal data to generate high-quality responses.

Even though RAG removes the requirement for a heavy-weight LLM for generation, retrieval still has a high overhead. 
The core of a RAG system is a vector database that enables vector similarity search. 
Different from LLMs, the overhead of RAG mainly comes from its memory footprint. For example, a Flat Index stores and sequentially searches every vector representation of the data chunks to identify the closest match to the query.
For example, in fever\cite{thorne-etal-2018-fever} dataset, a vector database that holds 5.23 million records has an index size of 18.5 GB. 
In comparison, mobile devices tend to have 4-12 GB of main memory \cite{Wiens_2024, Samsung_2024}.  Thus even using the whole memory on a mobile platform is not sufficient to run a large vector database. On the other hand, storing the vector database on disk introduces substantial access latency, impacting performance.

Our research focused on the challenges of implementing Retrieval Augmented Generation (RAG) on edge systems—mainly the overhead of the vector similarity search. 
We find that naively keeping the entire index in the main memory would not fit into the memory of mobile platforms. 
A Flat index which performs sequential search of all embeddings not only expensive in term of computation, but also trash memory leading to poor performance. 
On the other hand, Two-level Inverted File (IVF) index clusters embeddings of data chunks into centroids. The retrieval process first searches for the closest centroid and then performs a second search within the cluster.
Thus, avoid expensive sequential search of all embeddings.
However, keeping all embeddings in memory still leads to excessive memory thrashing and increased latency. Thus, keeping the first-level centroid in memory and generating the second level online can be a promising solution.
Through further profiling the RAG on mobile platforms using widely-used RAG benchmarks \cite{thakur2021beir}, we find that both the data access pattern and the access latency are highly skewed. 
First, most of the embedding vectors are not searched during the retrieval process. 
Second, the cost of generating the embedding vector is not the same for all clusters and has an extreme tail distribution.
These skewness leave space for further optimizations.

In this work, we develop a mobile RAG system, \xname{} that enables RAG-based LLM on mobile platforms, by fitting the vector database in the limited mobile memory while ensuring that the response time meets the service level objectives (SLOs) of mobile AI assistant applications. 
Based on these observations, our key ideas are the following: 
First, we prune the vector embedding of the data embeddings within centroid clusters which are only used for second-level search to save the memory capacity.
\xname{} then generates the vector embedding online during the retrieval process. However, due to limited computing on edge systems, generating vector embedding online could suffer from long embedding generation latency from large tail clusters. 
To overcome this challenge, our second solution is to pre-compute and store the embeddings of large tail clusters to avoid long tail latency of generating vector embeddings of data within those tail clusters. Then, \xname{} can adaptively cache the remaining vector embeddings to minimize redundant computation and improve overall latency, based on the spare memory capacity and SLO requirements. 

We evaluate \xname{} on a mobile platform based on Nvidia Jetson Orin Nano equipped with 8 GB of shared main memory,  similar to a mobile edge platform that has neural processing capabilities \cite{Samsung_2024}.
We use 6 workloads from the BEIR benchmark suite \cite{thakur2021beir} and tune the retrieval hyperparameters to normalize the recall against the Flat index baseline.
We also evaluate the generation quality using GPT-4o \cite{gpt4o} LLM as an LLM evaluator \cite{saad2023ares}. 
We use the time-to-first-token (TTFT) latency as the main metric. 
The result shows that \xname{} offers 1.8 $\times$ faster TTFT over the baseline IVF index on average and 3.82 $\times$ for larger datasets. At the same time, \xname{} maintains a similar generation quality with recall and generation scores within 5 percent of the Flat Index baseline while allowing all of our evaluated datasets to fit into the memory and avoid memory thrashing.

In summary, the contributions of this work are the following:
\begin{itemize}[leftmargin=*]
    \item We identify two key challenges of implementing RAG on edge devices: First, limited memory capacity does not allow loading large embedding vector database in the memory leading to memory thrashing and poor performance. Although only a small subset of embeddings are searched for two-level IVF index. Second, limited computing power of edge devices which slows down online embedding generation especially on few large tail and repeatedly used clusters.
    \item To enable scalable and memory-efficient RAG on edge systems, we propose \xname{} which improves upon IVF index by pruning second-level embeddings to reduce memory footprint and generate the embedding online during retrieval time. \xname{} mitigates long tail latency from generating the embeddings of heavy cluster by pre-computing and storing those tails. To further optimize latency, \xname{} selectively caches generated embeddings to reduce redundant computation while minimizing memory overhead.
    \item We implement \xname{} on Jetson Orin edge platform and evaluate our system with datasets from BEIR benchmark. The results show that \xname{} significantly improves the retrieval latency of large datasets with embedding sizes larger than the memory capacity by 131\% with slight reduction in retrieval and generation quality.
\end{itemize}
\section{Background}
\subsection{Retrieval Augmented Generation}\label{subsub:rag}
Retrieval Augmented Generation (RAG) \cite{lewis2020retrieval} is a technique used to improve the accuracy of the Large Language Model (LLM) by integrating an external knowledge base with the LLM. Instead of retraining the entire LLM with new information, RAG adds new data to the database. The RAG consists of two parts: Indexing and Lookup.

\begin{figure}[h!]
    \centering
    \begin{subfigure}[c]{0.95\columnwidth}
    \includegraphics[clip,width=\textwidth]{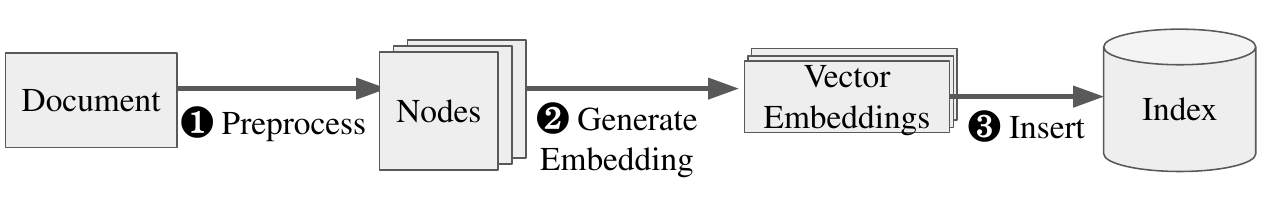}
    \caption{Indexing}
    \label{fig:rag_index}
    \end{subfigure}%
    \centering
    
    \begin{subfigure}[c]{0.95\columnwidth}
    \includegraphics[clip,width=\textwidth]{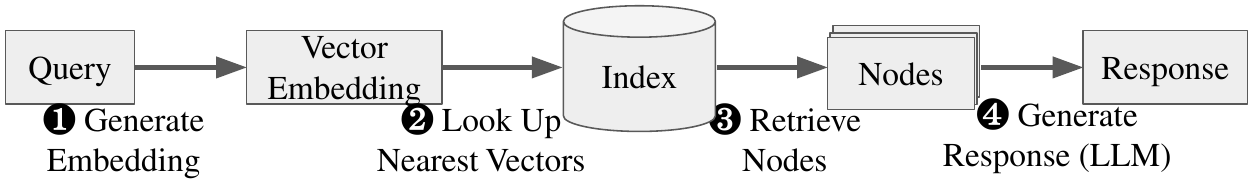}
    \caption{Lookup}
    \label{fig:rag_lookup}
    \end{subfigure}%
    \vspace{-0.3cm}
    \caption{RAG Pipelines.}
    \label{fig:rag_pipeline}
    \vspace{-1em}
\end{figure}

\textbf{Indexing:} Adding new data to the database involves a process called indexing, illustrated in Figure \ref{fig:rag_index}. First, the system pre-processes the incoming data. This preprocessing involves splitting the data into smaller overlapping chunks represented by nodes (step \encircle{1}). Each chunk of data is then fed into an embedding model, which generates a unique high-dimension vector representation for that particular piece of data (step \encircle{2}). Finally, these embedded vectors, which act as compressed representations of the data, are stored in a separate database for efficient retrieval later (step \encircle{3}).

\textbf{Lookup:} When a user submits a request, the system performs a lookup process as shown in Figure \ref{fig:rag_lookup}.  First, the user's request goes through the same embedding model as the data in an indexing phase (step \encircle{1}). This creates a vector representing the user's query.  Then, this embedding of the query is compared to the embedded vectors stored in the dedicated database. This is also known as vector similarity search (step \encircle{2}). The system returns the closest matches among these stored indexes (step \encircle{3}). Finally, the data nodes associated with the closest matching indexes are retrieved and fed to the large language model (LLM) to generate a response (step \encircle{4}).  Essentially, the lookup process helps retrieve data in the database related to the user's query. The LLM then uses this retrieved information to generate a response.

\subsection{Index Similarity Search}
As mentioned in the previous section, RAG searches for the embedding representation of the data chunk in the database with the closest distance to the query embeddings. However, embeddings often reside in high-dimensional spaces. Techniques like sorting, which are efficient in low-dimensional spaces, become less effective or even infeasible in high-dimensional spaces. Consequently, finding similar embeddings in high-dimensional spaces requires computationally expensive operations. Distance metrics such as Euclidean distance or cosine similarity involve calculations across all dimensions of the vectors, making the process time-consuming.

\subsection{RAG Indexing Methods}
We previously discussed the challenges of vector similarity search. Now, we'll explore indexing techniques that can significantly improve search.

\textbf{Flat Index} compares a query embedding to every embedding in the index. While this method is highly accurate, it can be computationally expensive, especially for large datasets.

\begin{figure}
    \centering
    \includegraphics[width=1\linewidth]{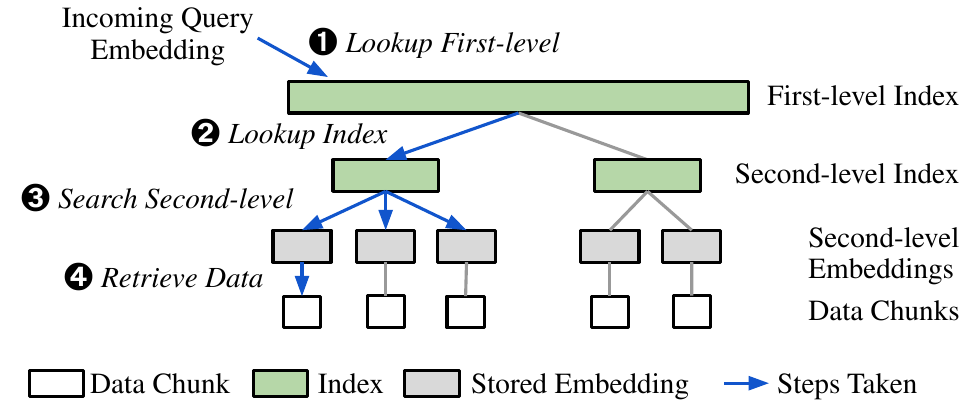}
    \caption{Retrieval process of Inverted File Index}
    \label{fig:tree_probe}
\end{figure}

\textbf{Inverted File (IVF) Index} \label{subsub:ivf}
\cite{ivf} employs clustering algorithms to group similar documents into clusters. Figure \ref{fig:tree_probe} illustrates IVF's retrieval procedure. During query processing, the system first compares the query to the centroids of these clusters in the first level index (step \encircle{1}). This initial comparison significantly reduces the number of documents to be examined, leading to faster search times. Then the system searches the second level index for the most similar embedding (step \encircle{2}). Finally, the system retrieves data associated with the embedding (step \encircle{3}). However, this approach can potentially miss relevant documents that don't align perfectly with cluster boundaries. To mitigate this, the system may also need to search adjacent clusters, which can slightly increase search time but improve accuracy.

\section{Motivation}\label{sec:motivation}
\subsection{Memory Limitation on Edge Platforms}

\begin{table}
\setlength{\tabcolsep}{1pt}
\small
    \centering
    \caption{Edge system comparison}
    \begin{tabular}{lll}
       \toprule
       System & Memory & Compute Units \\
       \midrule
       iPhone 16 Pro \cite{Wiens_2024} & 8 GB & CPU+GPU+NPU \\
        Galaxy S24 \cite{Samsung_2024} & 8 GB & CPU+GPU+NPU \\
       Jetson Orin Nano \cite{orinnano} & 8 GB & CPU+GPU+TensorCore \\
    \midrule
       Nvidia L40 \cite{L40} (Server) & 48 GB & GPU\\
       \bottomrule
    \end{tabular}
    \label{tab:sw_interface}
\end{table}

\begin{figure}
    \centering
    \includegraphics[width=1\linewidth]{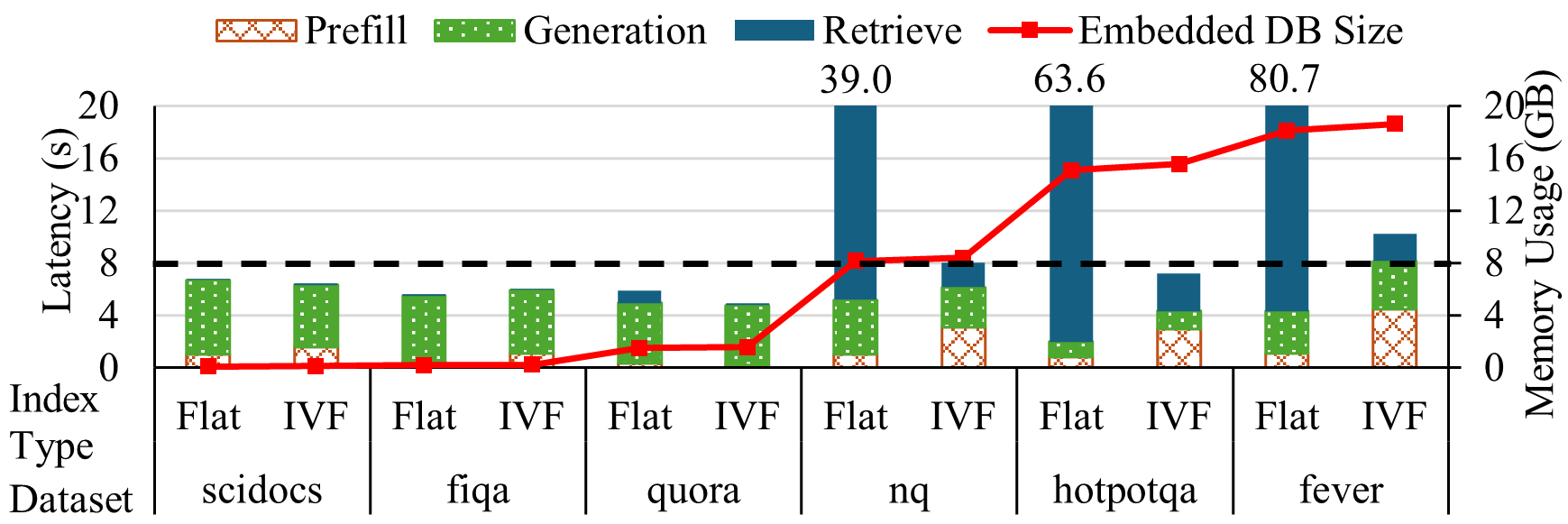}
    \caption{RAG latency breakdown and embedded database size}
    \label{fig:motivation_breakdown}
\end{figure}

Figure \ref{fig:motivation_breakdown} presents the latency breakdown of the RAG system on an edge platform. End-to-end latency is categorized into three phases: 1) Retrieval latency, the duration of vector similarity search to identify relevant text chunks; 2) First Token Latency or Prefill, the time from model input (query and retrieved chunks) to the generation of the first output token; and 3) Generation latency, the time from the first to the last generated token. The combined Retrieval and First Token latency, or Time-to-First-Token (TTFT), directly affects user-perceived latency, as it represents the delay between query submission and the display of the initial response.
While IVF index exhibits slower latency growth compared to Flat index, both configurations are susceptible to memory thrashing for datasets exceeding available memory (nq, hotpotqa, fever). This results in significant latency penalties as the system is forced to repeatedly load and unload the embedding database and the model from the storage. Therefore, efficient memory utilization is critical to mitigate this issue.

\subsection{Compute vs. Data Movement trade-offs}
\begin{figure}
    \centering
    \includegraphics[width=1\linewidth]{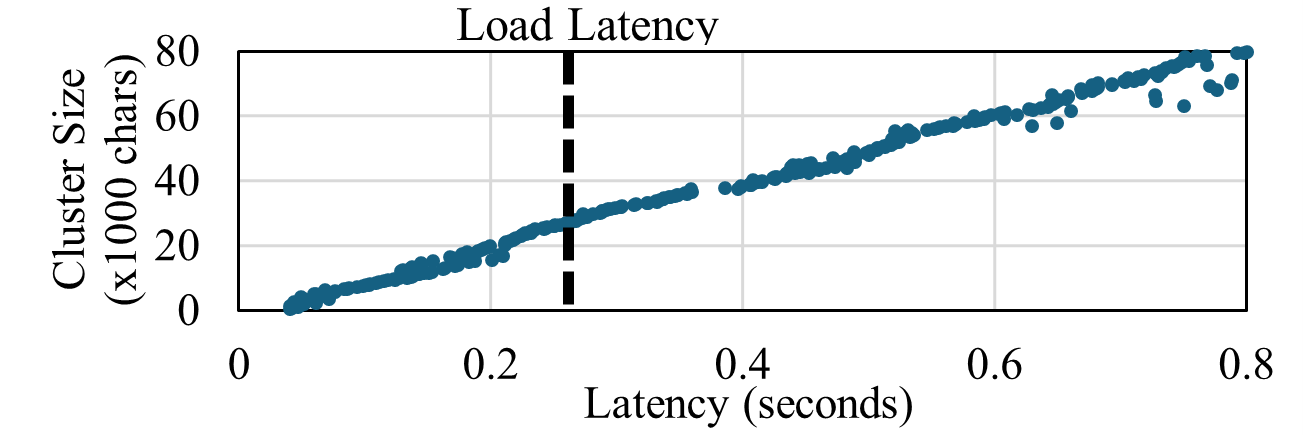}
    \caption{Embedding Generation Rate of different cluster size}
    \label{fig:motivation_gen_load}
\end{figure}

One of the way to reduce the memory footprint of embeddings is to generate them only when needed for incoming queries.
Since the two-level IVF index primarily searches embeddings linked to a small subset of first-level centroids, most of these embeddings can be generated during the retrieval process itself. This strategy avoids the significant memory overhead of storing all embeddings.
As shown in Figure \ref{fig:motivation_gen_load}, generating embeddings for clusters smaller than 24000 characters or approximately 8000 tokens is faster than loading them from the storage. This indicates that online embeddings generation can not only save memory but also potentially improve retrieval latency. However, generating embedding for large cluster could also cause long tail latency.

\section{High-level Ideas}\label{sec:keyidea}

We present \xname{}, a memory-efficient RAG system by selectively pruning second-level embeddings. Upon receiving an incoming query, \xname{} generates related second-level embeddings during runtime.  However, online embedding generation presents several key challenges:
(1) High Latency Variability: Embedding generation time is influenced by data chunk size, leading to unpredictable latency.
(2) High Computational Cost: Generating embeddings is computationally expensive.
Next, we describe the high-level ideas of \xname{} in this section.

\subsection{Selective Index Storage}\label{subsec:selective}
Pruning second-level embeddings reduces memory footprint, but shifts embedding generation from indexing to retrieval. This can increase retrieval latency if generating embeddings takes longer than accessing pre-computed ones. Here we make several observations related to embedding generation.

\begin{figure}
    \centering
    \includegraphics[width=1\linewidth]{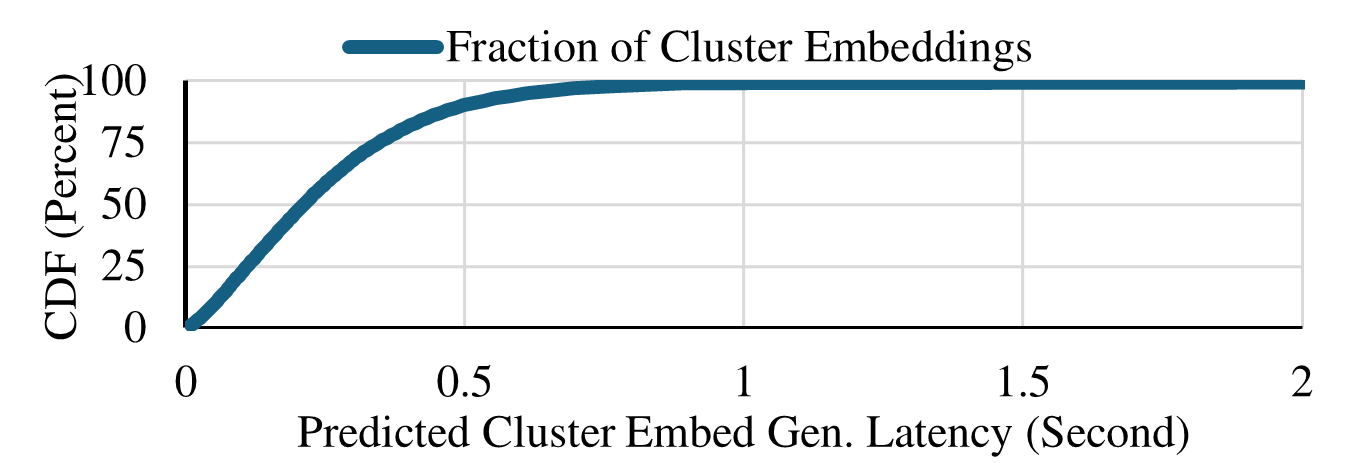}
    \caption{Cluster Embedding Generation Cost of nq dataset}
    \label{fig:embed_gen_cost}
\end{figure}

Figure \ref{fig:embed_gen_cost} presents the distribution of embedding generation times for various clusters within the nq datasets. The majority of clusters exhibit generation latencies under 500 milliseconds. Nevertheless, a subset of clusters, albeit infrequent, may experience generation times exceeding 2 seconds. This distribution highlights a "tail-heavy" characteristic, where a small proportion of clusters significantly impact the overall retrieval time.

\begin{figure}
    \centering
    \includegraphics[width=1\linewidth]{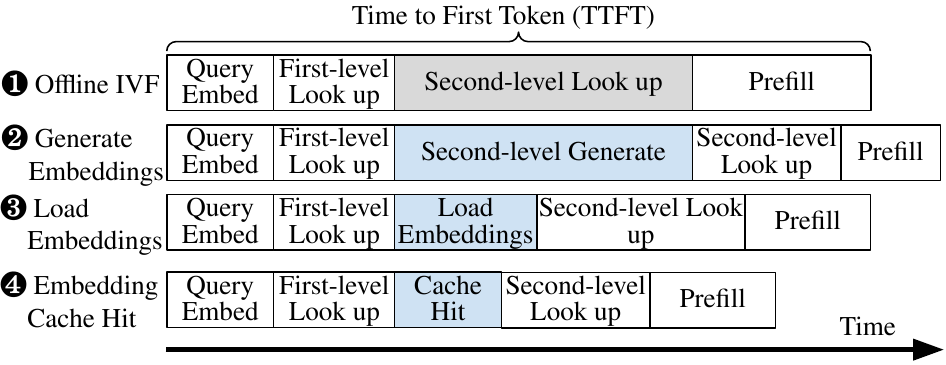}
    \caption{Timing breakdown of \xname{} Retrieval Process.}
    \label{fig:timeline}
\end{figure}

To mitigate long-tail latency in large clusters, \xname{} employs a hybrid approach. During the initial indexing phase, clusters are profiled to estimate embedding generation latency. Clusters exceeding the latency threshold are identified and their embeddings are precomputed and stored. At query time, embeddings are retrieved from storage if available (\encircle{3} in Figure \ref{fig:timeline}), bypassing the long latency of online embeddings generation phase (\encircle{2} in Figure \ref{fig:timeline}). For clusters not precomputed, embeddings are generated in flight. Algorithm \ref{alg:index} illustrates our Selective Index Storage.

\begin{algorithm}[tb]
   \caption{Selective Index Storage}
   \label{alg:index}
\begin{algorithmic}
\small
\STATE DataEmbeddings = embed(Datachunks)
\STATE Centroids = cluster(Embeddings)

\FOR{DataEmbedding in DataEmbeddings}
    \STATE ${\rm Centroid} = {\rm Centroids.search}({\rm DataEmbedding})$
    \STATE Centroid.add(DataEmbedding)
\ENDFOR
\FOR{Centroid in Centroids}
    \STATE // Compute Cluster's Embedding Gen. Latency\\
    ${\rm Centroid.GenLatency}$\\=${\rm SUM}({\rm len}({\rm Centroid.Datachunks})) / {\rm GenRate}$
    \IF{${\rm Centroid.GenLatency} > {\rm SLO}$}
        \STATE ${\rm Centroid.saveEmbeddings}()$
    \ENDIF
\ENDFOR
\end{algorithmic}
\end{algorithm}

\subsection{Adaptive Cost-Aware Caching}\label{subsec:cache}

Beyond the latency challenges presented by generating large cluster embeddings, we notice frequent reuse for embeddings of smaller, more common clusters. An analysis of queries across various datasets revealed a substantial degree of overlap in the accessed clusters. This overlap is quantified in Table \ref{tab:datasets} through the chunk reuse ratio. The significant reuse observed across all datasets implies that embeddings for clusters associated with these frequently accessed data chunks must be repeatedly generated. Therefore, \xname{} also caches generated embeddings to avoid redundant computation. Figure \ref{fig:timeline} shows the process of cluster embedding cache hit (\encircle{4}).

\begin{table*}
\setlength{\tabcolsep}{3.5pt}
\small
    \centering
    \caption{Evaluated datasets.}
    \begin{tabular}{lccccccc}
       \toprule
       Dataset & Corpus & \# Records & Embeddings & Unique Access & Total Access & Reuse Ratio & Fit in Dev. Mem \\
       \midrule
       scidocs \cite{specter2020cohan} & 86 MB & 3.6 k & 113 MB & 1157 & 2000 & 1.73 & \cmark \\
       fiqa \cite{www18}& 130 MB & 25 k & 217 MB & 2974 & 13286 & 4.47 & \cmark \\
       quora \cite{Quora}& 641 MB & 523 k & 1.5 GB & 15672 & 30000 & 1.91 & \cmark \\
       nq \cite{kwiatkowski-etal-2019-natural} & 4.6 GB & 2.68 M & 8.3 GB & 8186 & 10235 & 1.25 & \xmark\\
       hotpotqa \cite{thorne-etal-2018-fever}& 11 GB & 5.42 M & 15.4 GB & 15519 & 22098 & 1.42 & \xmark \\
       fever \cite{yang-etal-2018-hotpotqa}& 7.5 GB & 5.23 M & 18.5 GB & 5783 & 13922 & 2.41 & \xmark \\
       \bottomrule
    \end{tabular}
    \label{tab:datasets}
\end{table*}

To optimize cache utilization, \xname{} strategically avoids caching embeddings from smaller clusters with lower generation latencies. This approach balances cache hit rates and overall latency. Caching all embeddings can lead to low hit rates and high latency, while exclusively caching expensive embeddings can improve hit rates but increase overall latency. A carefully selected threshold prevents the caching of low-cost embeddings, striking a balance between the two.

To achieve this, \xname{} evicts and prevents caching of cluster embeddings whose generation latency falls below a dynamically adjusted Minimum Latency Caching Threshold. Algorithm \ref{alg:threshold} details how this threshold is adjusted. Initially, the threshold is set to 0, effectively caching all cluster embeddings. Subsequently, \xname{} gradually increases the threshold while continuously monitoring cache hit rates and the moving average of retrieval latency. If a cache miss occurs, and the current retrieval latency is lower than the moving average, the threshold is further increased. Conversely, if a cache miss does not occur, the threshold is decreased. This adaptive mechanism ensures that the cache prioritizes embeddings from clusters with significant generation costs to improve the overall performance gains.

\begin{algorithm}[tb]
   \caption{Cost-aware Least-Frequently Used Replacement Policy}
   \label{alg:replacement}
\begin{algorithmic}
\small
\STATE Incoming cache access with cluster index i
\STATE cluster = Cache.search(i)
\IF{exist(cluster)}
    \STATE ${\rm cluster.counter}++$
\ELSE
    \STATE // Get Weighted Least Frequently Used Cluster
    \STATE ${\rm minCost} = {\rm MAXVALUE}, {\rm evictClusterIndex} = -1$
    \FOR{cluster in Cache}
        \IF{${\rm cluster.genLatency} \times {\rm cluster.counter} < {\rm maxCost}$}
        \STATE ${\rm minCost} = {\rm cluster.genLatency} \times {\rm cluster.counter}$
        \STATE ${\rm evictClusterIndex} = {\rm cluster.index}$
        \ENDIF
    \ENDFOR
    \STATE ${\rm cache.delete}({\rm evictClusterIndex})$
    \STATE // Get data chunks and generate embeddings
    \STATE ${\rm dataChunks} = {\rm getDataChunks}(i)$
    \STATE ${\rm embeddings} = {\rm embed}({\rm dataChunks})$
    \STATE ${\rm cache.insert}({\rm index} = i, {\rm embeddings})$
\ENDIF
\STATE // Update Counters
\FOR{cluster in Cache}
    \STATE ${\rm cluster.counter} = {\rm cluster.counter} \times {\rm decayFactor}$
\ENDFOR
\end{algorithmic}
\end{algorithm}

\begin{algorithm}[tb]
   \caption{Minimum Latency Caching Threshold}
   \label{alg:threshold}
\begin{algorithmic}
\small
\STATE ${\rm MinimumLatencyCachingThreshold} = 0$ // Initialize
\FOR{each Query}
    \IF{embedCacheMiss = True} 
        \IF{${\rm movAvgLatency} < {\rm lastLatency}$}
            \STATE ${\rm MinimumLatencyCachingThreshold}++$
        \ENDIF
    \ELSE
        \STATE ${\rm MinimumLatencyCachingThreshold}--$
    \ENDIF
\STATE ${\rm movAvgLatency} = (1-\alpha) \times {\rm movAvgLatency} + \alpha \times {\rm lastLatency}$
\ENDFOR
\end{algorithmic}
\end{algorithm}

\begin{figure}
    \centering
    \includegraphics[width=1\linewidth]{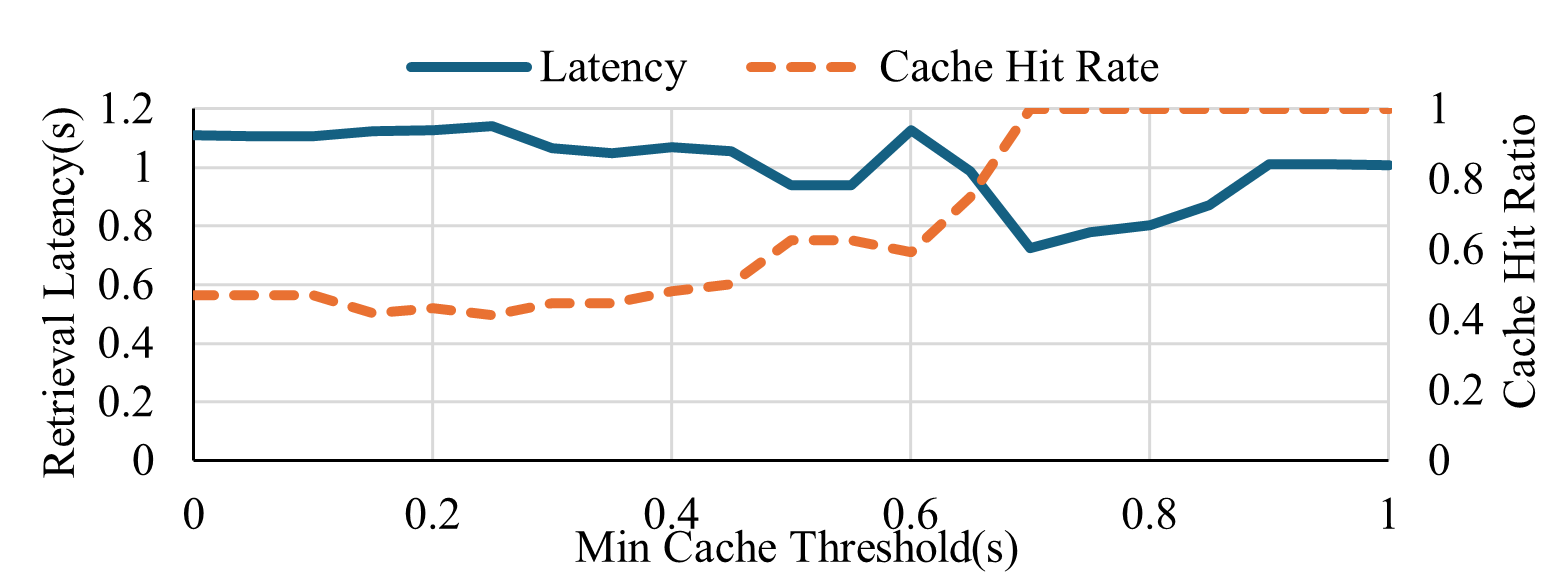}
    \caption{Retrieval Latency and Cache Hit rate with different Minimum Caching Threshold of fever dataset.}
    \label{fig:cache_threshold}
\end{figure}

\section{\xname{} System}
\subsection{Overview}
\xname{} index is a two-level indexing system that exploits memory efficiency and online computation. \xname{} index is based on the traditional two-level Inverted File (IVF) Index. The first level, always residing in memory, stores cluster centroids and references to the second level index. The second level stores the references to the text chunks, and the embedding generation latency of all data chunks. However, instead of storing all text chunk embeddings, the embeddings are pruned and \xname{} generates them online during the retrieval process, only indexes of costly clusters are stored (see Section \ref{subsec:selective}). To optimize performance and reduce latency, \xname{} employs a selective caching strategy for embeddings generated during the retrieval process (see Section \ref{subsec:cache}). The system prioritizes caching expensive embeddings, where cache hits yield significant performance gains. 
Embeddings that can be regenerated quickly without compromising service-level objectives (SLOs) are avoided to make room for those more expensive embeddings.

\subsection{\xname{} Indexing}
\begin{figure}
    \centering
    \includegraphics[width=1\linewidth]{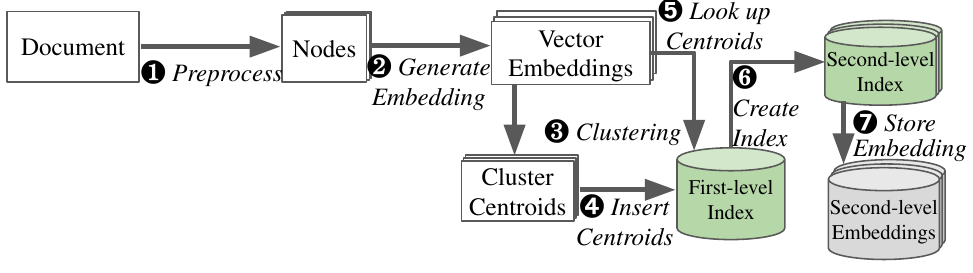}
    \caption{\xname{} Indexing Process}
    \label{fig:edgerag_indexing}
\end{figure}

\xname{} employs an indexing process similar to the Inverted File Index (IVF). Figure \ref{fig:edgerag_indexing} illustrates \xname{} indexing process. Initially, the text corpus is segmented into smaller data chunks (step \encircle{1}), and embeddings are generated for each chunk (step \encircle{2}). These embeddings are then clustered (step \encircle{3}), and the clusters' centroids are stored in the first-level index (step \encircle{4}). Then associated data chunks' embeddings are assigned to their respective clusters (step \encircle{5}) and references to the corresponding data chunks are stored (step \encircle{6}). Unlike traditional IVF, where all data embeddings are retained, \xname{} calculates the computational cost of generating embeddings for each data chunk within a cluster. If this cost exceeds a predefined threshold (the Service Level Objective, or SLO), the embeddings of the whole data chunk are stored (step \encircle{7}). Otherwise, the embeddings are discarded to optimize storage.

\subsection{\xname{} Retrieval}
\begin{figure}
    \centering
    \includegraphics[width=1\linewidth]{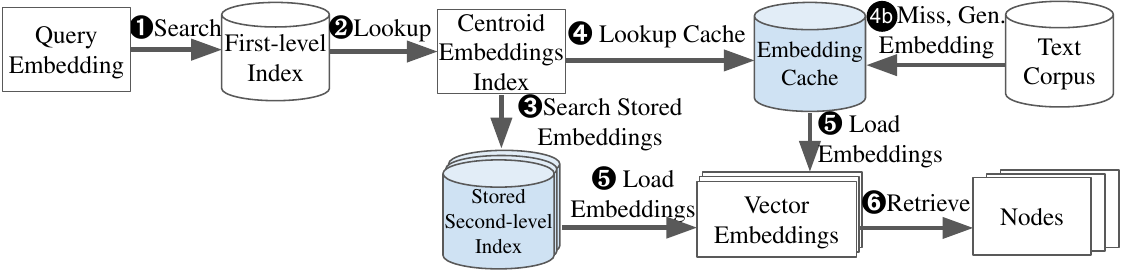}
    \caption{\xname{} Retrieval Process}
    \label{fig:edgerag_retrieval}
\end{figure}

As mentioned in Section \ref{sec:keyidea}, \xname{} incorporates both heavy cluster embedding loading and embedding caching.
Figure \ref{fig:edgerag_retrieval} shows \xname{} retrieval process. \xname{} first identifies the most similar centroid cluster to the query embedding (Step \encircle{1}). It then checks if pre-computed embeddings exist for this cluster (Step \encircle{2}). If so, \xname{} searches for stored embeddings (Step \encircle{3}) and these embeddings are loaded (Step \encircle{5}). If pre-computed embeddings are unavailable, \xname{} looks up the embedding cache (Step \encircle{4}). If the cache \emph{hits}, \xname{} loads the embedding from the cache and retrieves related data chunks (Step~\encircle{5}). If the cache \emph{misses}, \xname{} retrieves all associated data chunks in the cluster, regenerates the embeddings, and updates the cache (Step \encircle{4b}) before loading the embedding (Step \encircle{5}). \xname{} then looks up for the closest matching embedding and retrieves associated data chunks (Step \encircle{6}).

\subsection{\xname{} Insertion and Removal}
The \xname{} insertion process mirrors the initial indexing process. However, instead of clustering newly added embeddings, \xname{} identifies the existing cluster with the nearest centroid embedding. The index associated with this cluster is then updated. If the computational cost of the updated cluster's embeddings exceeds the SLO, \xname{} regenerates and stores its embeddings. In extreme cases where a cluster becomes excessively large, it is split into smaller clusters, and the newly created cluster is added to the first-level index. To remove data chunks, \xname{} first locates the corresponding cluster, then removes the associated embeddings, and the cluster index is subsequently updated. If the computational latency of generating the embedding of the cluster falls below the Service Level Objective (SLO), the entire cluster's embedding can be eliminated. In cases where a cluster becomes too small, it may be merged with a neighboring cluster. This removal process can be performed asynchronously, as leaving the cluster small has no immediate adverse impact on retrieval latency.
\section{Evaluation}
\subsection{System Setup}
We evaluate \xname{} on an Nvidia Jetson Orin Nano evaluation kit. We leverage Llamaindex \cite{Liu_LlamaIndex_2022} as our RAG framework and FAISS \cite{douze2024faiss} as our vector store. NanoLLM \cite{nanollm} powers our LLM engine, utilizing Sheared\_LLaMA-2.7B \cite{xia2023sheared} for generation. We employ gte-base-en-v1.5 \cite{li2023gte} as our embedding model. Table \ref{tab:eval_platform} provides a detailed overview of our evaluation setup and associated system software.

\begin{table}[t]
\setlength{\tabcolsep}{3pt}
    \centering
    \caption{Evaluation Platform. }
    \small
    \begin{tabular}{ll}
        \toprule
        \multicolumn{2}{c}{\textbf{Evaluation Hardware}} \\
        \midrule
        CPU &  Cortex A78AE, 1.2 GHz, 6-core\\
        GPU &  Ampere, 1024 CUDA cores, \\
        & 32 Tensor cores, 625 MHz \\
        Power Limit & 15 W\\
        DRAM & 8 GiB LPDDR5-4250\\
        Storage & 512 GB SD Card, UHS-I\\
        \midrule
        \multicolumn{2}{c}{\textbf{Software Configuration}} \\
        \midrule
        RAG Framework & LlamaIndex v0.11.18\\
        Vector Store Engine & FAISS 1.7.4\\
        LLM Engine & NanoLLM 24.6\\
        Embedding Model & gte-base-en-v1.5 (Dim=768)\\
        Generation Model & Sheared-LLaMA-2.7B\\
        System Software & Jetpack 6.0\\
        \bottomrule
    \end{tabular}
    \label{tab:eval_platform}
\end{table}

\subsection{Methodology}
We evaluate \xname{} against five distinct configurations: a linear-search-based flat index, a two-level IVF index with precomputed cluster embeddings, a two-level index with online cluster embedding generation, a two-level index with online cluster embedding generation and large cluster loading from storage, and a two-level index with online cluster embedding generation, large cluster loading from disk, and caching (\xname{}). Table \ref{tab:index_config} shows different evaluated Index configurations. For configurations utilizing the two-level IVF index, the embedding clustering process, performed using FAISS K-means with 20 iterations, is pre-computed and shared across all four configurations.
\begin{table}[h]
\setlength{\tabcolsep}{2pt}
\centering
    \small
    \caption{Evaluated Index Configurations. }
    \label{tab:index_config}
    \begin{tabular}{lll}
        \toprule
        \multirow{ 2}{*}{Index configuration} &\multicolumn{2}{c}{Embeddings location} \\ 
        \cmidrule{2-3}
        & Level 1& Level 2\\ 
        \midrule
        Flat & Memory & N/A\\
        IVF & Memory & Memory\\
        IVF+Embed. Gen. & Memory & -\\
        IVF+Embed. Gen.+Load & Memory & Storage (Partial)\\
        \xname{} & Memory & Storage + Memory\\
        \bottomrule
    \end{tabular}
    \vspace{-1em}
\end{table}

To address the known precision and recall trade-offs inherent to IVF-based indexing methods \cite{ivf}, we optimize the retrieval hyperparameters, specifically the number of cluster probes and retrieved data chunks. This optimization is aimed at normalizing the recall metric to match that of the flat index baseline \cite{msmarco}.
We evaluate \xname{} on six datasets from the BEIR benchmark suite \cite{thakur2021beir}. Table \ref{tab:datasets} provides a breakdown of the corpus size, number of data records, and total embedding size for each dataset. Note that three of the workloads: nq, hotpotqa, and fever have embedding footprints larger than the memory capacity of our platform. Therefore, it is impossible for the whole embedding database to fit into the memory. We set the retrieval SLO similar to those in LLM serving systems \cite{kakolyris2024slo,aminabadi2022deepspeed}: 1 second for smaller \emph{scidocs}, \emph{fiqa} and \emph{quora} datasets and 1.5 seconds for larger \emph{nq}, \emph{hotpotqa} and \emph{fever} datasets.

\subsection{Results}

\subsubsection{Retrieval Quality Evaluation} \label{subsub:retrieval_eval}
\begin{figure}
    \centering
    \includegraphics[width=1\linewidth]{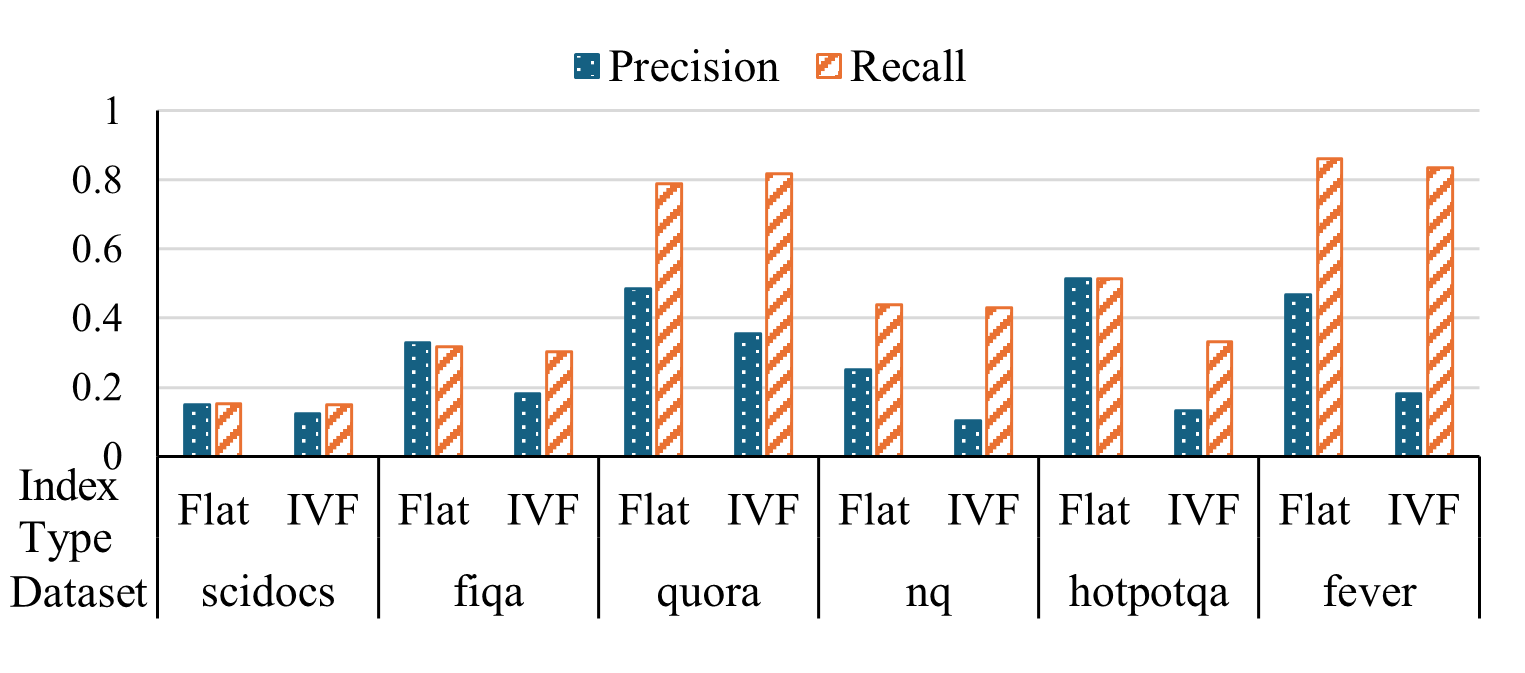}
    \caption{BEIR Evaluation Scores}
    \label{fig:beir_bench}
\end{figure}

First, we evaluate the retrieval quality using the BEIR benchmark. We concentrate on Flat and two-level IVF indexes, as \xname{}, which optimizes the IVF index, produces identical retrieval results to the two-level IVF index.
Figure \ref{fig:beir_bench} presents the precision and recall of different workloads. The results demonstrate a trade-off between precision and recall in two-level indexing schemes. While increasing the number of retrieved data chunks improves recall by including more relevant data, it also introduces more irrelevant data chunks, leading to a decrease in precision.

\subsubsection{Generation Quality Evaluation}
\begin{figure}
    \centering
    \includegraphics[width=1\linewidth]{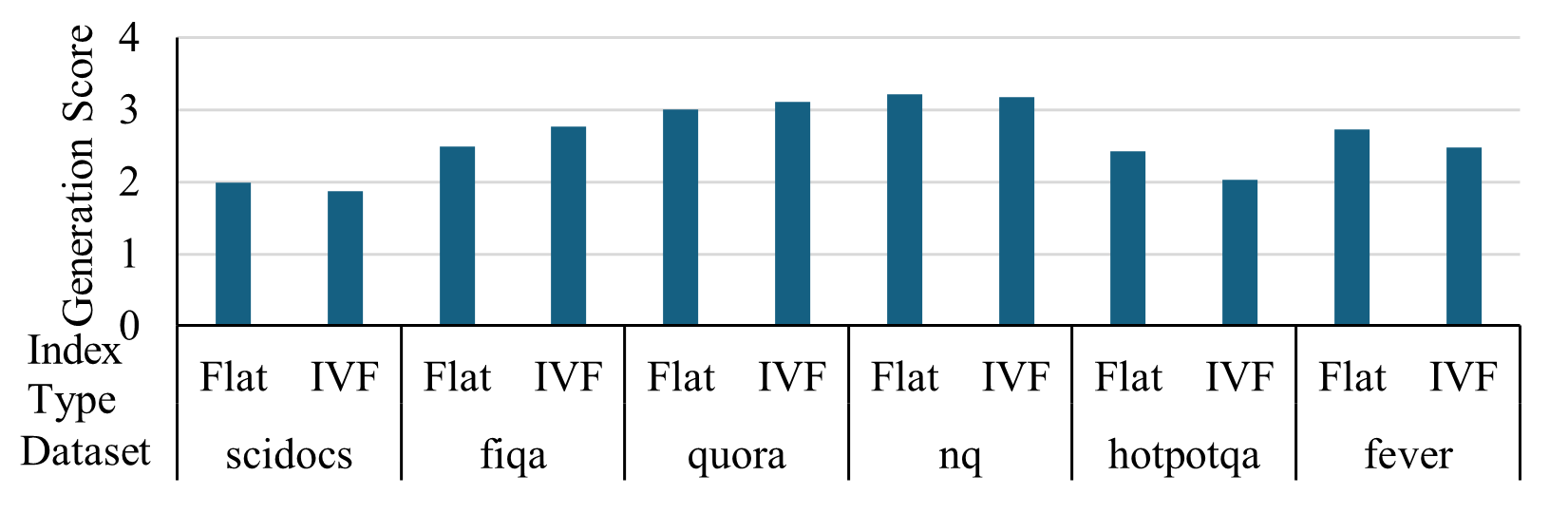}
    \caption{LLM Generation Evaluation Scores.}
    \label{fig:llm_eval}
\end{figure}

As mentioned in the previous section, IVF-based \xname{} exploits a precision-latency trade-off, prioritizing recall over precision. To investigate the potential benefits of prioritizing recall, we compare IVF-based \xname{} to a Flat index baseline. We employ a GPT-4o LLM \cite{gpt4o} as an evaluator to assess generation quality across various datasets. As shown in Figure \ref{fig:llm_eval}, while two-level IVF indexing schemes may exhibit lower precision, they can still achieve high-quality generation. This suggests that the generation model is capable of filtering out irrelevant information and leveraging only the most pertinent details for output generation.

\subsubsection{Retrieval Tail Latency Analysis}
\begin{figure}
    \centering
    \includegraphics[width=1\linewidth]{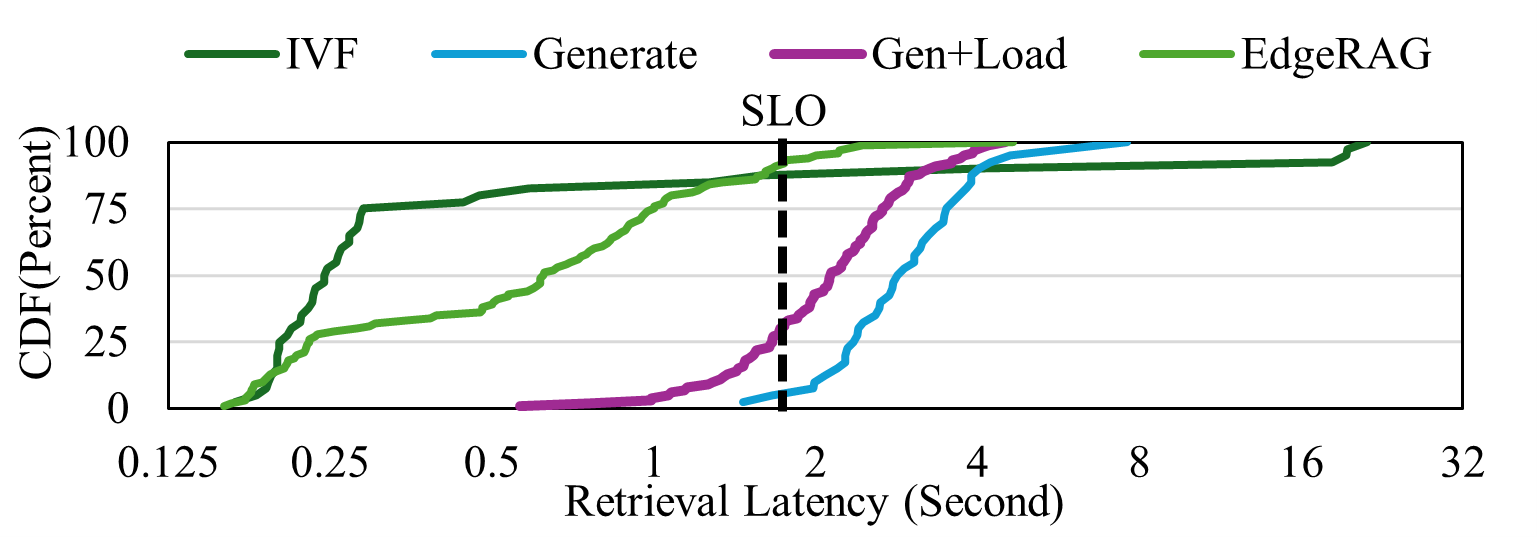}
    \caption{Retrieval Latency distribution with different optimization.}
    \label{fig:opt}
\end{figure}
In this section, we evaluate the tail latency of different \xname{}'s optimizations.
Figure \ref{fig:opt} illustrates the latency distribution of \xname{}'s various optimizations for nq dataset. The baseline IVF index, while achieving low latency for most queries, exhibits an extreme long tail latency, with the 95th percentile exceeding the median by over 64$\times$. This behavior stems from memory thrashing when accessing paged-out cluster embeddings.

The embedding generation configuration significantly improves latency by eliminating memory thrashing through embedding pruning. This results in a more than 4$\times$ reduction in 95th percentile latency compared to the IVF baseline. Further optimizing with large cluster loading from storage yields another 2$\times$ tail latency reduction. Finally, caching eliminates redundant embedding computation, leading to significant overall latency reduction. While \xname{} prioritizes resource efficiency to meet retrieval SLOs, allocating more resources such as allocating more cache capacity could unlock even greater latency improvement.

\subsubsection{End-to-End Latency}
\begin{figure}
    \centering
    \includegraphics[width=1\linewidth]{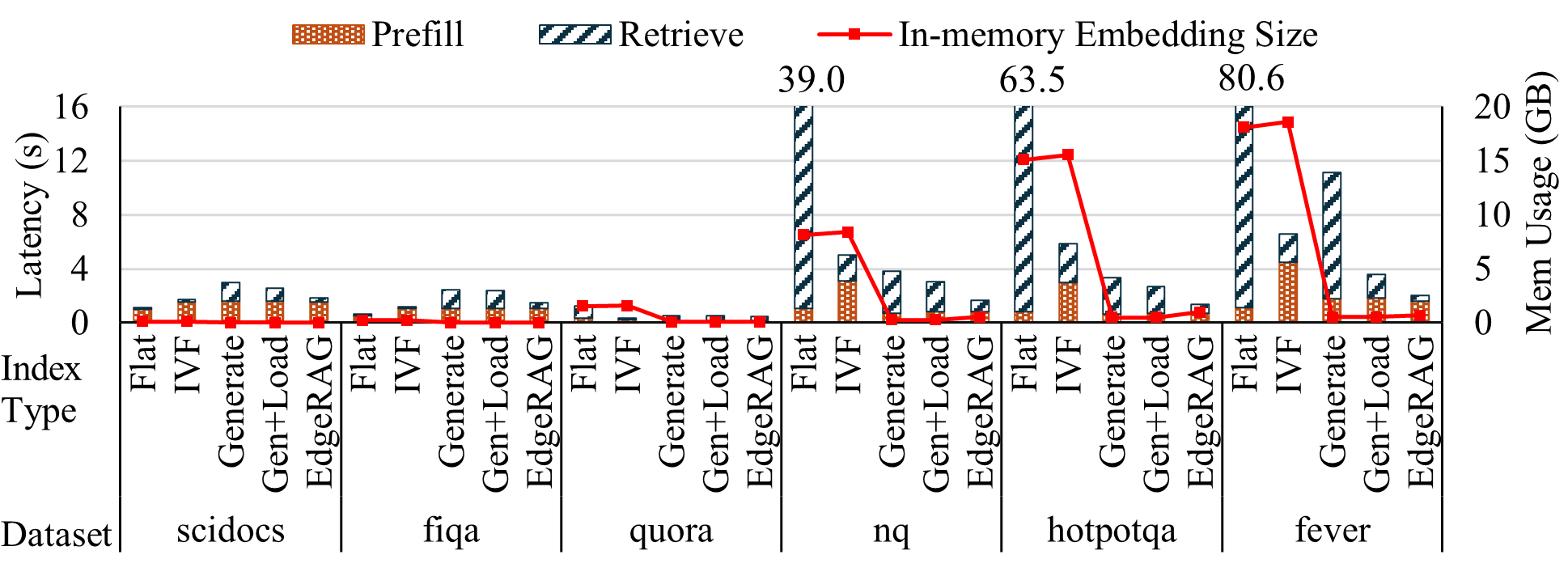}
    \caption{Retrieval and First Token Latency.}
    \label{fig:result_latency}
\end{figure}
In this section, we evaluate the end-to-end latency of a query. We use Time-to-First-Token (TTFT) as an evaluation metric. The TTFT consists of retrieval latency which includes vector similarity search, and prefill which is the time LLM takes to generate the first output token after it receives the prompt. We exclude Total Generation Time and Time-Per-Output-Token metrics as they are determined by the LLM's decoding rate, which is not optimized by \xname{}.

We then compare the latency of all five configurations (Table \ref{tab:index_config}). For the two smallest datasets: scidocs and fiqa, where the entire embedding dataset can fit into memory, the online embedding generation scheme without caching exhibits higher retrieval latency compared to both Flat and baseline pre-generated Two-level indexes. This is because in-memory operations, even linear searches, can be executed very efficiently in this scenario. For the Quora dataset, where the total embedding size nearly exceeds memory capacity, we observe that increasing the number of embeddings significantly degrades the performance of the Flat Index due to the overhead of linear search operations. For large datasets like nq, hotpotqa, and fever, where the total embedding size significantly exceeds memory capacity, the baseline two-level IVF index suffers from increased retrieval latency and first token generation latency. This is due to the combined effects of memory thrashing during large index access and the eviction of the generation model from memory, leading to slower search and longer generation times.

In contrast, the \xname{} index, with and without heavy cluster loading and caching, remains within memory constraints and maintains low first token generation latency. However, for datasets with imbalanced, tail-heavy, and large clusters like Fever, the online cluster embedding generation process can introduce significant latency. Persisting heavy-tail cluster embeddings and employing caching can mitigate this issue by reducing tail latency and avoiding redundant embedding generation. This significantly improves overall retrieval latency 1.8 $\times$ on average while caching only utilizes an additional 7\,\% of system memory on top of a two-level online embedding generation scheme.

\section{Discussion}
\paragraph*{\textbf{Limited Memory Capacity of Edge Systems}}
While some modern edge devices, such as upcoming mobile devices \cite{OnePlus13}, boast substantial memory capacity, memory capacity still remains a major bottleneck for supporting large-scale datasets.
\xname{} efficiently prunes the majority of second-level embeddings, enabling effective utilization of large memory capacities to support larger datasets with more embedding footprints or more powerful LLMs for enhanced response quality.

\paragraph*{\textbf{Integration with other RAG systems}}
In this work, we demonstrate \xname's effectiveness on a basic RAG system using text corpus data. While more advanced RAG systems may employ more sophisticated retrieval techniques or support diverse data types (multimodal), those techniques still rely on vector similarity search \cite{chen2022murag, lin2022retrieval}. As such, these systems can also benefit from \xname's optimizations.

\paragraph*{\textbf{Exploiting Hardware Accelerator}}
Modern edge devices commonly integrate multiple hardware accelerators within their System on Chip \cite{Gupta_2021, IPhone16}. One of those is the Neural Processing Unit (NPU) which enables efficient processing of deep learning tasks.
\xname{} can leverage these NPUs in several ways: 1. Enhance embedding generation throughput by harnessing the NPU's high throughput. 2. Offloading the embedding model's processing to the NPU---frees up the GPU or CPU for other tasks. This enables pipelining or parallelization of operations such as embedding generation, vector similarity search, and LLM prefill.
\section{Conclusion}
In this work, we propose \xname{}, a novel RAG system designed to address the memory limitations of edge platforms. \xname{} optimizes the two-level IVF index by pruning unnecessary second-level embeddings, selectively storing or regenerating them during execution, and caching generated embeddings to minimize redundant computations. This approach enables efficient RAG applications on datasets that exceed available memory, while preserving low retrieval latency and without compromising generation quality. Our evaluation results show that \xname{} improves retrieval latency by 1.22$\times$ on average and by a substantial 3.69$\times$ for large datasets that cannot fit in the memory.

\bibliography{bibshort/app}
\bibliographystyle{icml2025}

\end{document}